# Survey of modern Fault Diagnosis methods in networks

ZiJian Yang
Beijing Graduate School, Chinese Academy of Sciences
Beijing, China
youngzjam@126.com

Yong Wang
Beijing Graduate School, Chinese Academy of Sciences
Beijing, China
wangyong@gucas.ac.cn

*Abstract*—With the advent of modern computer networks, fault diagnosis has been a focus of research activity. This paper reviews the history of fault diagnosis in networks and discusses the main methods in information gathering section, information analyzing section and diagnosing and revolving section of fault diagnosis in networks. Emphasis will be placed upon knowledge-based methods with discussing the advantages and shortcomings of the different methods. The survey is concluded with a description of some open problems.

## I. INTRODUCTION

Fault diagnosis in networks comes from the equipment fault diagnosis, and was first presented in the 1960s. In 1967, under the push of the NASA, the ONR (Office of Naval Research) took charge of the management of the MFPG (Mechanical Failure Preventing Group) [1]. With the large and complex systems increasing in spaceflight, manufacture, navigation, nuclear industry and hospital, more faults appear. There are too many assemblies in each large and complex system to cooperate together well all the time. So faults are unavoidable and become serious problems that we must face. Since the advent of the computer networks, more and more application systems lie on networks to share knowledge and achieve greater efficiency in production. The reliability of networks has increasingly been an important issue. With the help of other fields, fault diagnosis in networks develops rapidly from 1970s.

In early time, fault diagnosis in networks depended on the professional knowledge and implements. According to the ISO/OSI model, one can use a network-tester to monitor and measure the parameters of networks on the lower three layers (Physical, Data Link, and Network layer); use the protocol- analyzer on all layers except for the Physical layer to find the network topology, capture and analyze data packages, collect and manage information about networks. One can also apply some testing commands on the Data-Link layer to test if it is broken such as "ping" and "traceroute".

This paper is organized as follows: section 2 outlines the modern fault diagnosis methods in three phases. Section 3 focuses on the knowledge-based method. Moreover different methods are displayed and compared. Some conclusions are drawn in the last section concerning the future direction of fault diagnosis in networks.

## II. MODERN FAULT DIAGNOSIS METHODS IN NETWORKS

Early fault diagnosis techniques are too simplex to find complex faults and rely too much on the professional experience. Compared with the rapid network developing in scale and amount, early fault diagnosis techniques are poor on collecting information, analyzing data, getting real root causes, and becoming inefficient.

Usually, the fault diagnosis in networks is plotted into three sections: information gathering, information analyzing, diagnosing and resolving [3] .This paper will discuss modern methods used in these sections according to the following order.

### A. Information gathering

There are four central protocols for managing in network fault diagnosis: the Internet Engineering Task Force Internet (IETF) defines the Simple Network Management Protocol (SNMP) [4]; International Organization for Standardization (ISO) defines the Common Management Information Protocol (CMIP) [5]; Transaction Language 1 (TL1) is widely used to manage optical network (SONET) and broadband access infrastructure in North America [6]. SNMP gains dominant market share and becomes the real industrial standard because of the support of CISCO and other main network equipment manufacturers.

Information gathering can be divided into three kinds: active, passive, active-passive method [7] [8]. Most active methods of gathering information depend on SNMP. In SNMP polling model, agents running on the aim network element and a central controller running on a computer are necessary. Central controller sends request for aim status to the agents periodically. Some network management systems use this method, like the Open View system of HP Company, the Net View system of IBM Company and the Net Management system of SUN Company. For sending request to agents and agents echoing timely, networks cost mu bandwidth and time on transferring and computing.

Passive gathering SNMP Trap makes the controller monitor the SNMP Trap, without sending anything. So this method is real-time. However, Trap is carried by the User Data gram Protocol (UDP), which can not ensure the quality of transmission. So passive gathering SNMP Trap is easy to lose something important.

For the equipment which does not support the SNMP, topology linkage query [9], ICMP message parsing [10], syslog analyzing can be used. Some commercial software adapts these methods, such as the SPECTRUM system of the Cable-tron Company, and the New Web NMS system of the Advent Company.

### B. Information analyzing

Information analyzing is a process in which useful symptom is extracted from fault information, and fault is located, and isolated. It can be divided into two groups: exact inference and approximate inference. Exact inference has following methods: graph reduction, combinatorial optimization, poly tree propagation; approximate inference has following methods: method based on simulation, method based on searching, and transformation method. Transformation method is more important than others two. （Shown in Fig 1.）

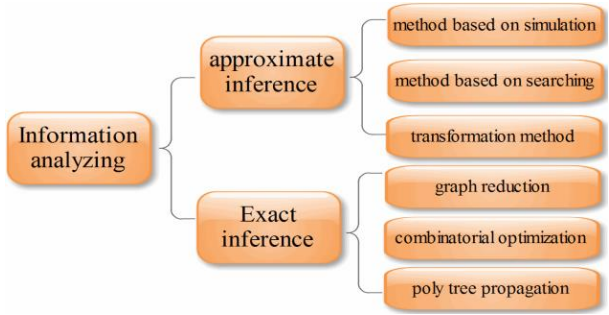

Figure 1. classification of methods on information analyzing

### C. Diagnosing and resolving

Diagnosis is a process that makes certain the location and type of fault. There are three classes: analytical model-based method, signal processing-based method and knowledge-based method [11]. Table I makes a comparison of them. (Shown in Table I)

TABLE I. METHODS IN DIAGNOSIS AND RESOLVING COMPARING

| method | technique adapted | advantage | disadvantage |
|---|---|---|---|
| signal processing-based method | extracting symptoms from fault information | achieve easily | misreporting and false alarm |
| analytical model-based method | state estimation with mathematical statistics , analytic functions | close to the truth | limitation to apply widely |
| knowledge-based method | expert system, fuzzy theory, fault tree, neural network, Bayesian network | intelligent and exact | limitation depend on the supported theory |

### III. OVERVIEW OF THE KNOWLEDGE-BASED METHOD

Among those methods which are used in the Diagnosis and resolving, the knowledge-based method becomes the primary research filed because of its self-rule and intelligence. The knowledge-based method is divided into many methods and technologies: fault diagnosis based on fault tree, fault diagnosis based on expert system, fault diagnosis based on Fuzzy Logic, fault diagnosis based on artificial neural network, fault diagnosis based on Grey theory, and fault diagnosis based on Bayesian networks. (Shown in Fig 2.)

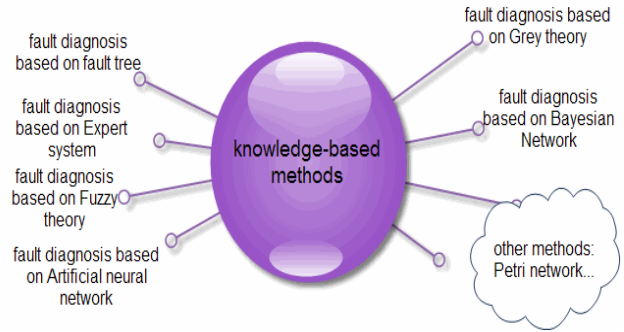

Figure 2. classification of knowledge-based methods

### A. Fault diagnosis based on fault tree (FT)

FT is a graphic deductive method which puts the worst fault status as "the top event". By searching all causes, FT names the cause which can arouse the fault directly as the second tier, the cause which can arouse the faults on the second tier directly as the third tier, and the basal cause as "the bottom event". All faults between the top event and the bottom event are named "the Intermediate event". FT looks for all possible fault models and gets the probability of the worst fault. It is an easy tool of fault diagnosis, but it is difficult to express the associated relationship, and less information content, limited self-educated ability and update ability. Researching on the combining FT with the neural network and expert system will be the future goal [12].

### B. Fault diagnosis based on Expert system(ES)

ES is the most remarkable achievement in fault diagnosis in recent years. It resolves problems with mimicking the behavior when human experts deal with these problems. ES is composed of knowledge base, inference engine, database, knowledge capturer, interpreter and human-machine interface. (Shown in Fig 3.)

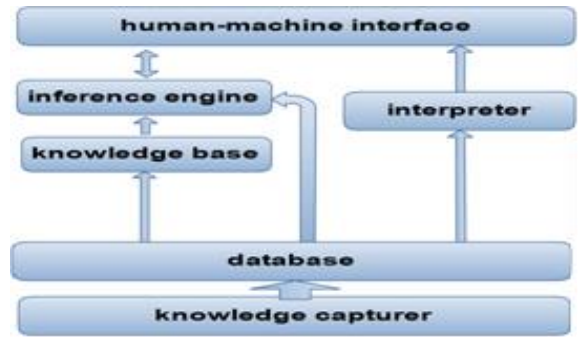

Figure 3. Relation between cells in expert system

The pivotal and difficult process of ES is knowledge capturing, knowledge expression and uncertainty reasoning [13] [14]. The fusion of the ES and fuzzy logic, FT, Artificial Neural Network will be the future of this researching field.

*C. Fault diagnosis based on Fuzzy Logic(FL)*

FL was proposed as the fuzzy set theory by Lotfi Zadeh in 1965. It is a form of many-valued logic and deals with reasoning that is approximate rather than fixed and exact.

Fault diagnosis based on FL depends on the mapping relationship between the symptom space and fault space, and reasons the fault from symptom. Because of the immature fuzzy theory, there is no uniform ways of how to measure the membership degree of element in fuzzy set theory and the mapping relationship between two fuzzy sets. Experience and vast examinations are usually the important ways to solve these problems [15] [16].

*D. Fault diagnosis based on artificial neura lnetwork(ANN)*

Artificial neural network (also neural network) is a mathematical and computational model that is inspired by the structure and functional aspects of biological neural networks. ANN consists of an interconnected group of artificial neurons, and it processes information using a connection approach to computation.

There are three researching fault diagnosis fields in which ANN is applied: as the classifier of diagnosis from the perspective of pattern recognition; as the dynamic prediction model to diagnosis from the perspective of prediction; as the builder of expert system based on ANN from the perspective of knowledge processing. The abilities of ANN, such as fault tolerant in principle, topology robustness, association, adaptive, self-learning, make it play an important role in complex system diagnosis, especially the multi-layer perception (MLP) theory based on back propagation algorithm (BP) is widely applied and successful. Although ANN has advantage in fault diagnosis of nonlinear system, it is non-robust at all. So the robustness algorithm and study on-line algorithm are the aim of ANN in the future. [17]

*E. Fault diagnosis based on Grey theory (GD)*

This method researches the relationship between information which is captured on the systemic point of view, i.e., detecting new, unknown diagnosis information from the known diagnosis information. It works on the Grey model, prediction and Grey correlation analysis. Because the Grey theory itself is incomplete, the Grey system diagnosis is limited with how to deduce the unknown information from the known things. [18]

*F. Fault diagnosis based on Bayesian Networks (BN)*

Bayesian networks ( also Belief networks or directed acyclic graphical model) is a probabilistic graphical model that represents a set of random variables and their conditional dependencies via a directed acyclic graph (DAG) and as one of the most effective models in the expression and reasoning of uncertain knowledge. Bayesian networks are directed acyclic graphs whose nodes represent random variables in the Bayesian sense: they may be observable quantities, latent variables, unknown parameters or hypotheses. Edges represent conditional dependencies; nodes which are not connected represent variables which are conditionally independent of each other. Each node is associated with a probability function that takes as input a particular set of values for the node's parent variables and gives the probability of the variable represented by the node.

Researchers have made progress in approximate inference. Stochastic sampling algorithm, search-based algorithm, model simplification algorithm and loopy belief propagation Search-based algorithm are improved in applicability, complexity, accuracy and efficiency by many researchers. However, none of the algorithms can be used widely; we must choose the best one according to the special problem. The compassion between some Bayesian networks reasoning algorithms is list in Table II. [19]

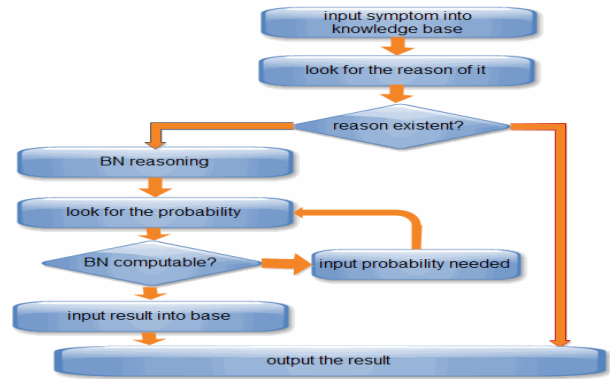

Figure 4. Structure of Bayesian networks diagnosis model

TABLE II. COMPASSION BETWEEN SOME BAYESIAN NETWORKS REASONING ALGORITHMS

| method | accuracy | key of algorithm | advantage |
|---|---|---|---|
| stochastic sampling | proportional to amount of sample | sampling | good result, wide application |
| search-based | depend on the selected state | right state, accurate state set | good Real-time computing |
| model simplification | depend on simplifing algorithm | simplify model, accurate estimation | simple, real-time reasoning |
| loopy belief propagation search-based | depend on iterative times | astringency of algorithm | good result with astringency |

## IV. CONCLUSION

The paper presents a survey of modern fault Diagnosis methods in computer networks, focuses on the contributions which we think close to the modern theory and may gain some relevance for the future research and practical applications.

As this paper expressed, fault diagnosis in networks has made great progress in common fault detecting and localization. Each method of fault diagnosis in networks relies on one or more theories, which determinates the application of method. In table III, the difference between the methods based on the knowledge is stated.

The fields which need to be strengthened are followed:
- Improving the gathering and analyzing ability in "soft fault", that means paying attention to the latent faults and symptoms. Gathering and analyzing them, drawing decision which can figure out problems before they appear.
- Improving the robustness of fault diagnosis algorithm.

TABLE III. DIFFERENCE BETWEEN THE METHODS BASED ON THE KNOWLEDGE

| method | knowledge-gathering | expression and inference | dependability of diagnosis | disadvantage |
| --- | --- | --- | --- | --- |
| fuzzy diagnosis | difficult | simple expression, difficult inference | depend on knowledge | easy to misdiagnosis |
| fault tree | difficult | simple expression, difficult inference | weak | short in uncertain status |
| expert system | difficult | difficult expression, difficult inference | depend on knowledge | poor uncertainty reasoning |
| artificial neural network | easy | difficult expression, simple inference | strong | short in exceptional fault |
| Bayesian networks | difficult | simple expression, simple inference | strong | consistency maintenance difficult |